\documentclass[sigconf]{acmart}
\copyrightyear{2025} 
\acmYear{2025} 
\setcopyright{none}
\acmConference[CIKM '25]{Proceedings of the 34th ACM International Conference on Information and Knowledge Management}{November 10--14, 2025}{Seoul, Republic of Korea} 
\acmBooktitle{Proceedings of the 34th ACM International Conference on Information and Knowledge Management (CIKM '25), November 10--14, 2025, Seoul, Republic of Korea} 
\acmDOI{10.1145/3746252.3760905} 
\acmISBN{979-8-4007-2040-6/2025/11}
\renewcommand\footnotetextcopyrightpermission[1]{}
\usepackage{multirow}
\usepackage{balance}

\begin{document}

\title{Solar Forecasting with Causality: A Graph-Transformer Approach to Spatiotemporal Dependencies}

\thanks{\copyright 2025 Copyright held by the owner/author(s). Publication rights licensed to ACM. This is the author’s version of the work. It is posted here for your personal use. Not for redistribution. The definitive Version of Record was published in CIKM '25. The definitive Version of Record will be available via ACM Digital Library at the DOI above.}

\author{Yanan Niu}
\authornote{Corresponding author}
\orcid{0000-0003-0589-9724}
\affiliation{%1
  \institution{EPFL}
  \city{Lausanne}
  %\state{Vaud}
  \country{Switzerland}
}
\email{yanan.niu@epfl.ch}

\author{Demetri Psaltis}
\orcid{0000-0003-4684-8800}
\affiliation{%
  \institution{EPFL}
  \city{Lausanne}
  %\state{Vaud}
  \country{Switzerland}
}
\email{demetri.psaltis@epfl.ch}

\author{Christophe Moser}
\orcid{0000-0002-2078-0273}
\affiliation{%
  \institution{EPFL}
  \city{Lausanne}
  %\state{Vaud}
  \country{Switzerland}
}
\email{christophe.moser@epfl.ch}

\author{Luisa Lambertini}
\orcid{0000-0001-6425-4606}
\additionalaffiliation{%
  \institution{Università della Svizzera italiana (USI)}
  \city{Lugano}
  \country{Switzerland}
}
\affiliation{%
  \institution{EPFL}
  \city{Lausanne}
  %\state{Vaud}
  \country{Switzerland}
}
\email{luisa.lambertini@epfl.ch}

\renewcommand{\shortauthors}{Yanan Niu, Demetri Psaltis, Christophe Moser, and Luisa Lambertini}
%% No italics, no superscripts
%% Use footnote or author note to identify equal contribution and/or contact author info

\begin{abstract}
Accurate solar forecasting underpins effective renewable energy management. We present \textbf{SolarCAST}, a causally informed model predicting future global horizontal irradiance (GHI) at a target site using only historical GHI from site $X$ and nearby stations $S$---unlike prior work that relies on sky-camera or satellite imagery requiring specialized hardware and heavy preprocessing. To deliver high accuracy with only public sensor data, SolarCAST models three classes of confounding factors behind $X$–$S$ correlations using scalable neural components: (i) observable synchronous variables (e.g., time of day, station identity), handled via an embedding module; (ii) latent synchronous factors (e.g., regional weather patterns), captured by a spatio-temporal graph neural network; and (iii) time-lagged influences (e.g., cloud movement across stations), modeled with a gated transformer that learns temporal shifts. It outperforms leading time-series and multimodal baselines across diverse geographical conditions, and achieves a 25.9\% error reduction over the top commercial forecaster, \textit{Solcast}. \textbf{SolarCAST} offers a lightweight, practical, and generalizable solution for localized solar forecasting.\footnote{Code available at \url{https://github.com/YananNiu/SolarCAST}}

\end{abstract}



\maketitle
\noindent\textbf{ACM Reference Format.}  
\noindent Yanan Niu, Demetri Psaltis, Christophe Moser, and Luisa Lambertini. 2025. Solar Forecasting with Causality: A Graph-Transformer Approach to Spatiotemporal Dependencies. \textit{In Proceedings of the 34th ACM International Conference on Information and Knowledge Management (CIKM ’25), November 10–14, 2025, Seoul, Republic of Korea.} ACM, New York, NY, USA, 5 pages. \url{https://doi.org/10.1145/3746252.3760905}

\section{Introduction}
By the end of 2024, global solar photovoltaic capacity surpassed 2.2 terawatts, providing over 10\% of the world’s electricity and accounting for more than 75\% of new renewable generation~\cite{IEA2025}. As solar power becomes a cornerstone of modern energy systems, precise short-term forecasting of solar irradiance is critical for grid stability, energy storage optimization, and energy market operations.

Traditional solar forecasting models typically rely on historical global horizontal irradiance (GHI) measurements at a single target site. However, these point-based models often struggle under dynamic weather conditions---particularly when clouds move rapidly across the sky. To address this, researchers have explored auxiliary inputs such as all-sky images~\cite{Gao2022}, panoramic cameras~\cite{yanan2025}, or satellite imagery~\cite{si2021satellite}. While these sources can enhance short-term predictions by providing broader spatial context, many depend on specialized hardware or remote sensing access. Also, they offer only indirect visual cues about local irradiance, making them less reliable under poor image quality or limited coverage.

In this work, we explore a less-tapped yet increasingly abundant auxiliary input for localized solar forecasting: GHI measurements from nearby weather stations. These data are now widely available across many countries\footnote{For example, Switzerland’s MeteoSwiss \url{https://www.meteoswiss.admin.ch/} and the U.S. National Solar Radiation Database \url{https://nsrdb.nrel.gov}.}, offering structured numerical signals with explicit physical meaning. Compared to imagery-based proxies, these time-series inputs are easier to preprocess and integrate, avoiding the need for complex image processing pipelines. Combining the target location’s GHI sequence $X$ with auxiliary sequences $S$ from nearby stations yields a spatio-temporal (ST) multivariate time series (MTS). A forecasting model is then applied to predict future GHI values at the target site, framing the task as an MTS prediction problem with a single target output.

To better leverage ST solar data, we adopt a causal perspective: while $S$ serves as auxiliary input for forecasting future GHI at the target site $X$, both may be influenced by shared confounders. We identify three key types: (1) Observable synchronous confounders (e.g., time of day, node identity) that affect all sensors simultaneously; (2) Latent synchronous confounders, such as regional weather patterns that create shared but unobserved influences across locations; and (3) Time-lagged causal drivers, like moving clouds, which introduce delayed, directional dependencies across space and time.
%(1) Synchronous confounders---including observable factors (e.g., time of day, seasonality) and latent regional conditions---affect all sensors at the same time, creating shared signals across locations. (2) Time-lagged causal drivers, such as rapidly moving clouds, introduce delayed, directional dependencies across time and space, influencing upstream sensors before the target.

These causal patterns call for models that capture both synchronous and asynchronous ST interactions. For synchronous effects, spatio-temporal graph neural networks (STGNNs) offer a natural fit, and have demonstrated strong performance in domains like traffic forecasting~\cite{li2018_DCRNN,Yu2018_STGCN,zhao2019tgcn,guo2021_ASTGNN}, neuroscience~\cite{li2021braingnn}, and recommendation systems~\cite{ying2018recommending}. By representing sensor relationships as a graph, STGNNs allow each node to aggregate information from spatial neighbors at the same time step, effectively modeling shared environmental influences. However, most existing STGNNs treat space and time separately, typically applying recurrent neural networks (RNNs), convolutional neural networks (CNNs), or other temporal modules before or after graph-based message passing~\cite{gao2022equivalence, cini2023taming}. This “time-then-graph” structure models spatial and temporal dependencies separately. However, this separation limits its ability to capture asynchronous, non-local interactions---such as a moving cloud affecting one sensor at time $t$ and another at $t + \Delta$. Since these patterns require joint modeling of spatiotemporal causality, we propose \textbf{SolarCAST}, a \textbf{C}ausal-\textbf{A}ware \textbf{S}patio-\textbf{T}emporal predictor, designed to more effectively capture both synchronous and asynchronous cross-series dependencies:
%This “time-then-graph” design models spatial and temporal dependencies separately, which makes it less suited for capturing asynchronous, non-local interactions, such as a moving cloud affecting one sensor at time $t$ and another $t+\Delta$, where joint modeling of ST causality is needed. 
\begin{itemize}
    \item An \textbf{embedding layer} encodes observable synchronous confounders for $X$ and $S$, such as time and station identity.
    \item A \textbf{STGNN module} implicitly captures synchronous \emph{long-term, stable spatial dependencies} between $X$ and $S$.
    \item A \textbf{Gated Transformer module} explicitly models \emph{non-regular, non-local ST interactions}. Its self-attention mechanism captures arbitrary dependencies across time and space, enabling the model to track causal patterns with temporal shifts. A gating mechanism filters out spurious correlations, emphasizing high-impact, informative relationships.
\end{itemize}

\textbf{SolarCAST} is lightweight and scalable, making it well-suited for real-world deployment across diverse geographic regions, even with sparse or uneven sensor distributions. It outperforms strong MTS baselines and multimodal models using image-based inputs, demonstrating the effectiveness of nearby GHI as a practical data source for this task. Compared to a commercial solar forecasting provider, Solcast\footnote{\url{https://solcast.com/}}, \textbf{SolarCAST} achieves a 25.9\% improvement in localized solar forecasting accuracy, highlighting its practical value.

\section{Framework}

Let $ X = \{x_{t-T:t}\} \in \mathbb{R}^{T \times d} $ denote the historical features at the target node, i.e., GHI at the forecasting site over a window of $T$ time steps. Similarly, let $ S = \{s^i_{t-T:t}\}_{i=1}^{N-1} \in \mathbb{R}^{T \times d} $ represent the historical features from $N-1$ auxiliary nodes, corresponding to nearby GHI sensors, which may be spatially sparse or unevenly distributed. The objective is to predict the GHI at the target node at a future time step, $ x_{t+h} := y $, given a prediction horizon $ h $.

\begin{figure}
  \includegraphics[width=\linewidth]{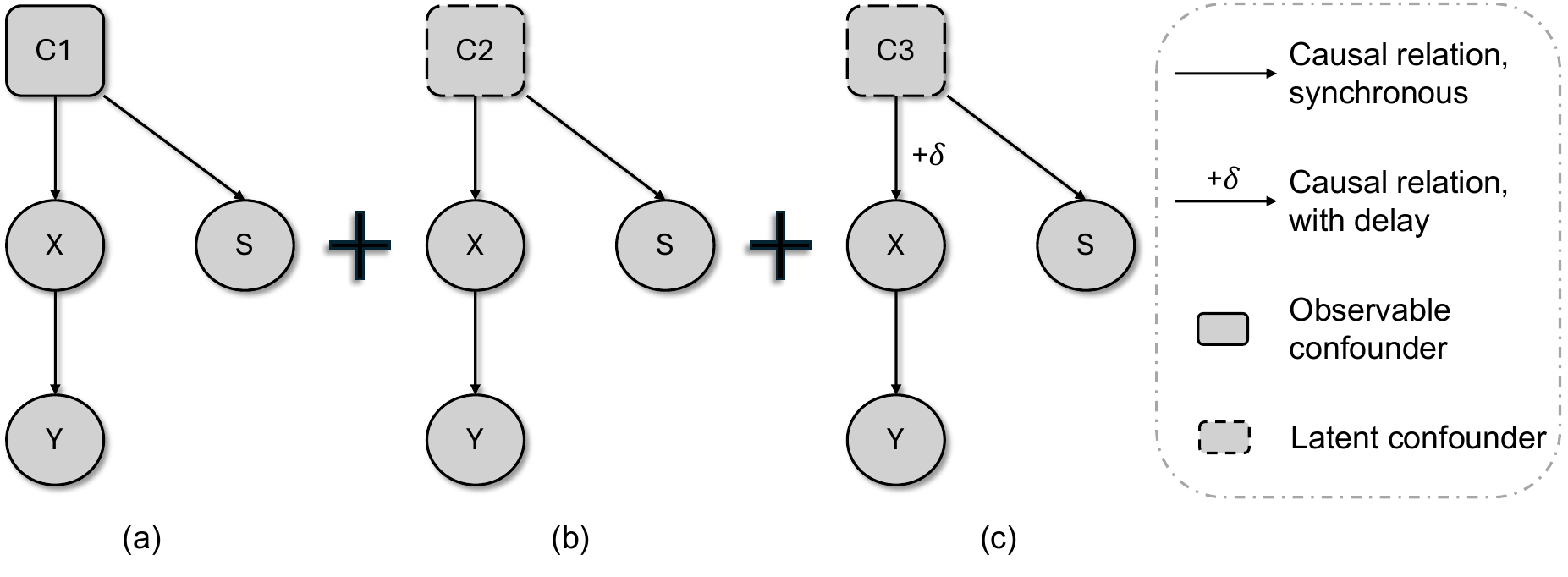}
  \caption{Illustration of the causal relationships underlying the correlation between the target and auxiliary sensors.}
  \Description{The illustration shows three types of confounders: observable synchronous confounders, latent synchronous confounders, and latent confounders with temporal delays.}
  \label{fig:causality}
  \vspace{-10pt}
\end{figure}

Previous MTS forecasting studies have adopted the Structural Causal Model (SCM) paradigm~\cite{Pearl2009} to improve model interpretability through a causal view---for example, by applying causal treatments~\cite{xia2023deciphering} or explicitly modeling confounders~\cite{Ji2025kdd}. Following this direction, we also frame our problem using SCM to guide the architectural design of our model and enhance forecasting accuracy. Specifically, we identify three types of confounders that may underlie the observed correlations between $X$ and $S$, see Fig.~\ref{fig:causality}:

\begin{itemize}
  \item \textbf{C1: Observable synchronous confounders}---such as time of day, season, or node identity---simultaneously influence both $X$ and $S$.
  \item \textbf{C2: Latent synchronized confounders}---such as regional cloud cover or large-scale atmospheric phenomena---affect all nodes simultaneously, but are not directly observed.
  \item \textbf{C3: Temporal-lagged latent confounders}---such as moving clouds---impact the auxiliary nodes $S$ first and then reach the target node $X$ after a delay $\delta$. 
\end{itemize}

\section{Model Design\label{sec:model}}
\begin{figure}[ht]
  \includegraphics[width=\linewidth]{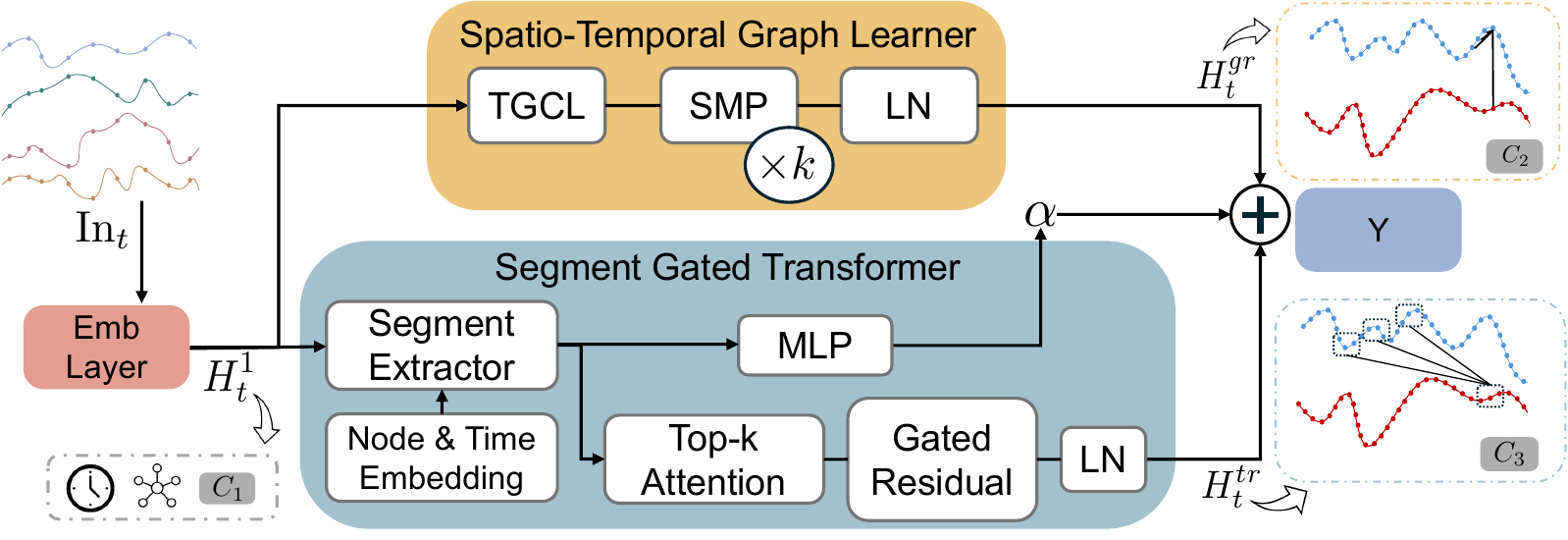}
  \caption{The pipeline of SolarCAST.}
  \Description{Three modules of SolarCAST are illustrated, each addressing a different type of confounder.}
  \label{fig:model}
  \vspace{-10pt}
\end{figure}
SolarCAST (Fig.~\ref{fig:model}) explicitly models each causal path identified in Fig.~\ref{fig:causality}, aligning its architecture with the physical constraints and statistical dependencies in ST solar data. As input, the model takes a sequence of $n$ historical time series, $\mathrm{In}_t:= (X_{t-T:t},S^1_{t-T:t},...,S^{(n-1)}_{t-T:t}) $. To account for observable synchronous confounders ($C_1$), an \textit{Embedding Layer} (EMB) augments each node’s feature representation with relevant contextual information. The resulting encoded features are then processed by two parallel modules: (1) a \textit{Spatio-Temporal Graph Learner} (STGL) to capture spatially synchronized patterns ($C_2$), and (2) a \textit{Segment Gated Transformer} (SGT) to model dynamic, lagged influences ($C_3$). The final output is computed as a learnable combination of the two pathways, weighted by a parameter $\alpha$:
\begin{subequations}
\begin{align}
    H_t^{(1)} &= \mathrm{EMB} \left( \mathrm{In}_t \right)\\
    H_t^{gr} &= \mathrm{STGL} \left( H_{t}^{(1)}  \right)\\
    H_t^{tr} &= \mathrm{SGT} \left( H_{ t}^{(1)} \right), \quad \text{for } k = 3, \ldots, K \\
   H_t^{out} &= \alpha * H_t^{tr} + (1 - \alpha) * H_t^{gr}
\end{align}
\end{subequations}
%This design enables SolarCAST to effectively disentangle different sources of correlation, aligning the architecture with the causal structure of the data.

\subsection{Embedding Layer\label{model:emb}}
To control for observable synchronous confounders for $\mathrm{In}_t \in \mathbb{R}^{n \times T \times d}$, we incorporate a \textit{time-of-day embedding} $E_{\mathrm{time}} \in \mathbb{R}^{T \times d_1}$ (capturing temporal context based on sampling frequency) and a \textit{node identity embedding} $E_{\mathrm{node}} \in \mathbb{R}^{n \times d_2}$ (via a learnable lookup table which assigns a unique vector to each node).These auxiliary embeddings are concatenated with the input features along the dimension axis:
\begin{equation}
    H_t^{(1)}  = \mathrm{Concat}(\mathrm{In}_t, E_{\mathrm{time}}, E_{\mathrm{node}}) \in \mathbb{R}^{n \times T \times (d + d_1 + d_2)}.
\end{equation}

\subsection{Spatio-Temporal Graph Learner}
To capture synchronous but unobserved confounders ($C_2$)  across sensor locations, we employ an STGL that propagates information via spatial message passing. A dynamic ST graph, $\mathcal{G}^t = (V, E,X_{t-T:t},\\S_{t-T:t})$, is defined, where $V$ is a set of $n$ sensor nodes and $E$ denotes the edges between them. The adjacency matrix $A \in \mathbb{R}^{n, n}$ : $\{(i,j):(i,j)\in E\}$ encodes node connectivity. 

Unlike fixed-topology domains (e.g., traffic, power grids), solar networks lack structural connectivity since spatial proximity doesn't guarantee predictive relevance due to local cloud dynamics. We adopt a continuous, data-driven adjacency formulation inspired by Wu et al.~\cite{wu2020dots} to learn soft, unidirectional dependencies, and avoid underfitting from static distance-based adjacency matrices or overfitting from fully end-to-end ones learned on small graphs:
\begin{align}
M_1 &= \tanh(E_1 W_1) \\
M_2 &= \tanh(E_2 W_2) \\
\mathcal{A} &= \sigma\left(M_1 M_2^{T} - M_2 M_1^{T} \right) 
\end{align}
where all $E_{\{\cdot\}}$ are node embedding look-up tables, and all $W_{\{\cdot\}}$ are trainable weight matrices. The sigmoid activation $\sigma()$ produces dense, data-driven connectivity while avoiding excessive sparsity.

An STGL consists of a Temporal Gated Convolutional Layer (TGCL) for local temporal encoding, a Spatial Message Passing (SMP) module repeated $K$ times (with $k = 3,\ldots,K+2$) to handle both inflow and outflow information through each node, and a Layer Normalization (LN) layer for stability:
\begin{subequations}
\begin{gather}
    H_t^{(2)}  = \mathrm{TGCL} \left( H_{t}^{(1)}  \right)\\
    H_t^{(k)} = \mathrm{SMP}\left( H_t^{(2)} , H_t^{(k-1)}, \mathcal{A} \right) 
+ \mathrm{SMP}\left( H_t^{(2)}, H_t^{(k-1)}, \mathcal{A}^T \right)\\
    H_t^{gr} = \mathrm{LN}(H_t^{(K+2)})
\end{gather}
\end{subequations}

\textbf{Temporal Gated Convolutional Layer.} TGCL has been shown to improve performance in sequence modeling tasks~\cite{oord2016PixelCNN,oord2016wavenet,wu2020dots}: 
\begin{equation}
    H_{t}^{(2)}  = tanh(H_{t}^{(1)}\ast W^f) \odot \sigma(H_{t}^{(1)} \ast W^g)
\end{equation}
where $\ast$ denotes the convolution operator with kernel size $k$, and $\odot$ indicates element-wise multiplication:

\textbf{Spatial Message Passing.} SMP adopts a graph convolutional network that performs multi-hop propagation over $K$ steps~\cite{wu2020dots}:
\begin{gather}
H_t^{(k)} = \beta\,H^{(2)}_t \;+\;(1 - \beta)\,\widetilde{A}\,H_t^{(k-1)}, \quad \text{for } k = 3, \ldots, K+2 \\
H_t^{gr} = \sum_{i=2}^{K+2} H_t^{(i)}\,W^{(i)} 
\end{gather}
where $\widetilde{\mathcal{A}} = \tilde{D}^{-1}(\mathcal{A} + I)$, $\tilde{D}_{ii} = 1 + \sum_j \mathcal{A}_{ij}$, and $\beta$ is a residual weight. The final output aggregates features from all propagation layers via an MLP for adaptive feature selection.

\begin{table}[b]
\vspace{-12pt}
\centering
\caption{Dataset summary.}
\vspace{-10pt}
\begin{tabular}{p{1.2cm}p{1.5cm}p{0.5cm}p{1.2cm}p{2.2cm}}%{m{1.5cm}m{1cm}m{1.5cm}m{0.9cm}m{1cm}}
\toprule
\textbf{Target} & \textbf{Range} & \textbf{\# S} & \textbf{Dis.(km)} & \textbf{Camera des.} \\ 
\midrule
Urban(OO) & 2021.12.7-2024.12.31 &  7 & 7.8-37.4 & $360^{\circ}$, clear  \\ 
Urban(SS) & 2023.07.05-2025.5.15 & 9 & 19.9-47.2 & $360^{\circ}$, noisy and small sky portion\\ 
Valley & 2022.07.28-2025.5.15 & 8 & 3.6-33.3 & $200^{\circ}$, foggy and small sky portion \\ 
Lake & 2023.04.10-2025.5.15 & 11 & 7.4-58.2 & $350^{\circ}$, strong water reflection \\ 
Mountain & 2022.05.14-2025.5.15 & 10 & 7.4-42.0 & $320^{\circ}$, foggy, snowy \\ 
\bottomrule
\end{tabular}

\vspace{-2pt}

\label{tab:data}
\end{table}
%\vspace{-10pt}

\subsection{Segment Gated Transformer}

\begin{table*}[ht]
\centering

\caption{Comparison with benchmark methods across five datasets from diverse geographic regions. Lower RMSE and RSE, and higher CORR indicate better performance. The best results are shown in \textbf{bold}, and the second-best are \underline{underlined}.}
\vspace{-10pt}
\resizebox{\textwidth}{!}{
\begin{tabular}{lllllllllllllllll}
\toprule
\multirow{2}{*}{\textbf{Model}} & \multirow{2}{*}{\textbf{Input}} & \multicolumn{3}{c}{\textbf{Urban (open space)}} & \multicolumn{3}{c}{\textbf{Urban (streetscape)}} & \multicolumn{3}{c}{\textbf{Valley}} & \multicolumn{3}{c}{\textbf{Lake}} & \multicolumn{3}{c}{\textbf{Mountain}} \\
\cmidrule(lr){3-5} \cmidrule(lr){6-8} \cmidrule(lr){9-11} \cmidrule(lr){12-14} \cmidrule(lr){15-17}
& & RMSE & RSE & CORR & RMSE & RSE & CORR & RMSE & RSE & CORR & RMSE & RSE & CORR & RMSE & RSE & CORR \\
\midrule
Persistent model~\cite{pedro2012}  &$X$ &100.039& 0.694 & 0.758 & 161.390 & 0.541 & 0.855 & 179.402& 0.585 & 0.818 & 140.514 & 0.490 & 0.879 & 152.097& 0.536& 0.858 \\
SMT~\cite{yanan2025} &$X,\mathrm{Img}$ & 74.228 & 0.515 & 0.858 & 152.368 & 0.511 & 0.865 & 123.201 & 0.402 & 0.917 & 130.209 & 0.454 & 0.894 & 142.725 & 0.503 & 0.866 \\
LSTNet~\cite{Lai2017} &$X,S$ &88.008 & 0.611 & 0.805 & 142.497& 0.478& 0.881 & 129.945 & 0.424 & 0.908 & 131.237& 0.457& 0.890 & 140.820& 0.496& 0.869 \\
STID~\cite{shao2022spatial}& $X,S$ & 76.409& 0.530& 0.849 & \textbf{123.126} & \textbf{0.413} & \textbf{0.912} & 125.114 & 0.408 & 0.916 & 121.359& 0.423& 0.907 & 130.510 & 0.460 & 0.890 \\
MTGNN~\cite{wu2020dots} &$X,S$ &\underline{73.664} & \underline{0.511} & \underline{0.860} & 127.283 & 0.427 & 0.905 & 121.660 & 0.397 & 0.920 & 120.534 & 0.420 & 0.908 & 126.186 & 0.444 & 0.897 \\
TimeGNN~\cite{xu2023timegnn} &$X,S$ & 78.492& 0.545& 0.842 & \underline{123.350} & \underline{0.413} & \underline{0.912} & \underline{120.741} & \underline{0.394} & \underline{0.919} & \underline{120.192}& \underline{0.419} & \underline{0.913} & \underline{125.766} & \underline{0.443} & \underline{0.897}\\

SolarCAST&$X,S$ & \textbf{70.787}  & \textbf{0.491} & \textbf{0.871} & 127.030 & 0.426 & 0.906 & \textbf{120.512} & \textbf{0.393} & \textbf{0.920} & \textbf{118.693} & \textbf{0.412} & \textbf{0.912} & \textbf{124.957} & \textbf{0.440} & \textbf{0.899} \\

\bottomrule
\end{tabular}
}
\label{tab:full}
\vspace{-7pt}
\end{table*}

We employ a transformer-based architecture augmented with cross-series attention to capture time-lagged and non-local dependencies across multiple time series. This mechanism allows the model to associate temporally distant segments and inter-series interactions, making it particularly effective for modeling dynamic propagation patterns, see Fig.~\ref{fig:model}. To improve robustness, we integrate a gating mechanism that selectively suppresses weak or noisy attention connections, highlighting the most informative ST dependencies. 

\textbf{Segment Extractor and Embeddings.} Given input $H^{(1)} \in \mathbb{R}^{n \times T \times d'}$, we extract a query patch from the target node over the most recent $q$ time steps, $H^{\mathrm{query}}_{x, t-q:t} \in \mathbb{R}^{1 \times (q \cdot d')}$. Additionally, we collect $m$ support patches of length $l$ from all nodes, defined as $H^{\mathrm{support}}_{t-r-p-l:t-r-p} \in \mathbb{R}^{(m \cdot n) \times (l \cdot d')}$ for $p = 0, \ldots, m-1$. A fixed temporal gap $r$ is enforced between the query and support segments to encourage the model to learn delayed dependencies between current and past information. Each segment is then projected into a representation of shape $\mathbb{R}^{1 \times d_{\text{model}}}$ via a linear layer. Temporal and structural context is incorporated for each patch using time embeddings and learnable node embeddings, similar to~\autoref{model:emb}.

\textbf{Top-k Attention \& Gated Residual.} Tokens are passed through a transformer block with two key modifications. First, we apply a Top-K sparse multi-head attention mechanism in which each query attends only to its top-K most relevant support tokens per head. This design focuses attention on the most salient historical patterns, especially those reflecting delayed confounding effects ($C_3$). Second, we apply a Gated Linear Unit~\cite{dauphin2017language,gehring2017convolutional} to the attention output before the residual connection. This gating re-weights feature channels, suppressing irrelevant signals and enhancing informative ones, thereby improving robustness without introducing extra sparsity.

\textbf{Learnable $\alpha$.} A learnable scalar $\alpha \in (0, 1)$ dynamically weights the relative contribution of SGT and STGL on a per-sample basis,
\begin{equation}
    \alpha = \sigma\left( W_1 \cdot \text{ReLU}(W_0 H^{\mathrm{query}}) \right)
\end{equation}

It is computed from the query patch representation of the target node using an MLP, allowing the model to infer the relevance of time-lagged factors based on the current node state.
%The resulting $\alpha \in (0, 1)$ weight STGL and SGT on a per-sample basis.  dynamically modulates the influence of delayed dependencies on a per-sample basis, enabling the model to flexibly integrate long-range context with short-term representations from the STGL pathway.
\section{Experiments}
\subsection{Experimental Setup}
\textbf{Baselines.}  
We compare against both solar forecasting and MTS models. For solar, we use the persistence model~\cite{pedro2012}, a simple physical baseline, and SMT~\cite{yanan2025}, a multimodal solar predictor. For MTS, we include LSTNet~\cite{Lai2017}, a hybrid deep learning model; STID~\cite{Zeng2022AreTE}, a linear ST method; MTGNN~\cite{wu2020dots}, an STGNN benchmark; and TimeGNN \cite{xu2023timegnn}, an STGNN that constructs graphs from timestamps to capture temporal dependencies. We also include a commercial forecasting solution from Solcast~\footnote{\url{https://solcast.com/}} for practical benchmarking.

\textbf{Data.}  
We collect ST GHI data from weather stations operated by MeteoSwiss\footnote{\url{https://www.meteosuisse.admin.ch}}. For each target location, nearby GHI measurements from surrounding stations serve as auxiliary time series. All data are sampled at 10-minute intervals. Tab.~\ref{tab:data} summarizes the datasets, including landform type, data range, number and distance of auxiliary sensors to the target, and camera metadata (field of view and viewing conditions) used by SMT~\cite{yanan2025}.

\textbf{Implementation.}  
SolarCAST is implemented in PyTorch 2.6.0 with time and node embeddings of dimension 10. The SGT module employs TGCL with kernel size 3 and SMP with depth $k=3$. SGT pathway parameters include query/support patch lengths $l=6$, $m=9$, gap $r=3$, 4 transformer heads and Top-5 attention. We use a 70/20/10 train/validation/test split with 2-hour prediction horizon at 10-minute resolution. Training employs MSE loss with AdamW optimizer (learning rate $1E-3$ → $3E-4$, polynomial decay factor 0.5, 20 epochs), and early stopping after 20 epochs of validation stagnation. Evaluation metrics include root mean square error (RMSE), root relative squared error (RSE), and correlation (CORR)~\cite{yanan2025}.

\subsection{Main Results}
Comparisons with baseline models are presented in Tab.~\ref{tab:full}. To ensure a fair evaluation against the multimodal model SMT, which incorporates both $X$ and $\mathrm{Img}$~\cite{yanan2025}, we restrict all testing to daylight hours when image data is available.\footnote{Consequently, our error metrics are not directly comparable to those from public datasets that include nighttime periods, where near-zero irradiance values can artificially deflate errors.} As shown in Tab.~\ref{tab:data}, the quality and forecasting relevance of image data vary substantially across datasets due to differences in local climate and camera setup. SMT performs best with high-quality, clear panoramic images, as seen in the "Urban (Open Space)" dataset, where its results are comparable to or better than models using $X$ and $S$. However, under less favorable conditions, the contribution of image data declines. In contrast, auxiliary GHI data ($S$) consistently serves as an effective substitute for local visual input, even with sparse or uneven sensor deployment. This enables accurate localized forecasting without the need for dense image coverage, demonstrating the practicality of using $S$ alongside $X$ in our framework.

Models with ST identification, including STID~\cite{shao2022spatial}, MTGNN~\cite{wu2020dots}, and TimeGNN~\cite{xu2023timegnn}, also perform strongly, leveraging topological priors through learned embeddings or explicit graphs to capture spatial correlations and suppress low-informative nodes. These methods align well with the nature of ST solar data.

SolarCAST builds on these strengths but is further tailored to ST solar forecasting by capturing time-lagged confounders that reflect delayed and dynamic weather effects. It achieves the best performance on 4 out of 5 datasets, demonstrating robustness across diverse geographic regions and sensor configurations. On a representative 12-day test set from March 2024, SolarCAST achieves an RMSE of 82.507, outperforming Solcast with an RMSE of 103.913, marking a 25.9\% improvement. This result underscores the practical value of SolarCAST for accurate, real-world solar forecasting.

\subsection{Ablations}

Ablations on "Urban (Open Space)" dataset, shown in Tab.~\ref{tab:ablation}, validate the effectiveness of SolarCAST’s three modules in addressing the corresponding types of confounders.

\begin{table}[ht]
\vspace{-5pt}
\centering
\caption{Ablation study.}
\vspace{-10pt}
\begin{tabular}{lllll}
\toprule
\textbf{Metrics} & \textbf{SolarCAST} & \textbf{w/o EMB} & \textbf{w/o STGL} & \textbf{w/o SGT} \\ 
\midrule
RMSE & 70.787 &  72.799 & 80.037 & 75.051 \\ 
RSE & 0.491 & 0.505 & 0.555 & 0.521 \\ 
CORR & 0.871 & 0.863 & 0.839 & 0.856  \\

% \hline
% \multirow{2}{*}{Idx}&\multicolumn{2}{c}{\textbf{Ablation}} & \multicolumn{3}{c}{\textbf{Performance}}\\
% \cmidrule(lr){2-3} \cmidrule(lr){4-6}
%  & ST graph  & transformer  & RMSE & RSE & CORR \\ 
% \midrule
% \raisebox{.5pt}{\textcircled{\raisebox{-.9pt} {1}}} &\checkmark & \checkmark   &  150.896 & 0.451 & 0.893 \\ 
% \raisebox{.5pt}{\textcircled{\raisebox{-.9pt} {2}}} & \checkmark & -   &  154.794 & 0.463 & 0.888 \\
% \raisebox{.5pt}{\textcircled{\raisebox{-.9pt} 3}} & - & \checkmark   &   160.350 & 0.479 & 0.882 \\
\bottomrule
\end{tabular}

\label{tab:ablation}
\vspace{-10pt}
\end{table}

\section{Conclusion}
In this paper, we introduce SolarCAST, a causally informed spatiotemporal model that addresses confounders in solar forecasting using only publicly available time-series data. Its superior performance, combined with strong model interpretability, demonstrates high accuracy and strong real-world applicability.

\begin{acks}
This work was supported by the EPFL Solutions for Sustainability Initiative (S4S) grant. The authors thank Solcast for providing access to the data used in this study.
\end{acks}

\bibliographystyle{ACM-Reference-Format}
\balance
\bibliography{Reference}
\end{document}